\documentclass{article}

\usepackage{arxiv}

\usepackage[utf8]{inputenc} 
\usepackage[T1]{fontenc}    
\usepackage{hyperref}       
\usepackage{url}            
\usepackage{booktabs}       
\usepackage{amsfonts}       
\usepackage{nicefrac}       
\usepackage{microtype}      
\usepackage{lipsum}		
\usepackage{graphicx}
\usepackage{natbib}
\usepackage{doi}

\usepackage{siunitx}
\usepackage[super]{nth}
\usepackage[symbol]{footmisc}
\usepackage{amsmath}
\usepackage{cleveref}

\title{COVID-19 Detection from Exhaled Breath}

\author{
    Nicol\`{o} Bellarmino \\
    Politecnico di Torino
    \And
    Riccardo Cantoro \\
    Politecnico di Torino
    \And
    Michele Castelluzzo \\
    NanoTech Analysis Srl
    \And
    Raffaele Correale \\
    NanoTech Analysis Srl
    \And
    Giovanni Squillero \\
    Politecnico di Torino
    \And
    Giorgio Bozzini \\
    Hospital ASST Lariana
    \And
    Francesco Castelletti \\
    Universit\`{a} degli Studi dell'Insubria
    \And
    Carla Ciricugno \\
    NanoTech Analysis Srl
    \And
    Daniela Dalla Gasperina \\
    Universit\`{a} degli Studi dell'Insubria
    \And
    Francesco Dentali \\
    Universit\`{a} degli Studi dell'Insubria
    \And
    Giovanni Poggialini \\
    Universit\`{a} degli Studi dell'Insubria
    \And
    Piergiorgio Salerno \\
    Universit\`{a} degli Studi dell'Insubria
    \And
    Stefano Taborelli \\
    Universit\`{a} degli Studi dell'Insubria
}

\begin{document}
\maketitle

\begin{abstract}
    The SARS-CoV-2 coronavirus emerged in 2019, causing a COVID-19 pandemic that resulted in 7 million deaths out of 770 million reported cases over the next four years. The global health emergency called for unprecedented efforts to monitor and reduce the rate of infection, pushing the study of new diagnostic methods. In this paper, we introduce a cheap, fast, and non-invasive detection system, which exploits only the exhaled breath. Specifically, provided an air sample, the mass spectra in the 10--351 mass-to-charge range are measured using an original nano-sampling device coupled with a high-precision spectrometer; then, the raw spectra are processed by custom software algorithms; the clean and augmented data are eventually classified using state-of-the-art machine-learning algorithms. An uncontrolled clinical trial was conducted between 2021 and 2022 on some 300 subjects who were concerned about being infected, either due to exhibiting symptoms or having quite recently recovered from illness. Despite the simplicity of use, our system showed a performance comparable to the traditional poly\-merase-chain-reaction and antigen testing in identifying cases of COVID-19 (that is, 0.95 accuracy, 0.94  recall, 0.96 specificity, and 0.92 $F_1$-score). In light of these outcomes, we think that the proposed system holds the potential for substantial contributions to routine screenings and expedited responses during future epidemics, as it yields results comparable to state-of-the-art methods, providing them in a more rapid and less invasive manner.
\end{abstract}

%
%
\thispagestyle{empty}

\section*{Introduction}


The World Health Organization (WHO) reports that the COVID-19 outbreak spread all over the world in 2019 caused more than 770 million infections and 7 million deaths\footnote{\url{https://covid19.who.int/} (retrieved on \today)}. During the global health emergency from January 2020 to May 2023, unprecedented efforts have been made to monitor and reduce the rate of infection, including social restriction \cite{Chu2020,Vandenberg2021}.

Real-time quantitative polymerase chain reaction (RT-qPCR) has been extensively utilized for the detection of infected individuals and the management of illnesses. This technology enables the identification of SARS-CoV\nobreakdash-2 ribonucleic acid in nasopharyngeal or oropharyngeal swab samples \cite{Feng2020, Eissa2021}. Nevertheless, due to its high sensitivity, obtaining reliable results necessitates a robust experimental design, as well as a thorough understanding of normalization procedures \cite{rtqPCR}. False negatives, wherein infected individuals are incorrectly identified as healthy, may occur for various reasons, including technical variables in sample collection, transportation, and viral RNA handling, genetic variations, sample types, viral load, and the duration of viral exposure \cite{Bahreini2020}. Additionally, the need for authorized laboratories with at least Biosafety Level 2 (BSL-2) certification can lead to a strain on laboratory resources and potential delays in test result processing and reporting. This is further exacerbated by the costly equipment and reagents required \cite{Wang2020}.

Various alternative strategies have been suggested to offer tests that are swift, cost-effective, user-friendly, and capable of detecting infections at early stages. Exhaled breath contains respiratory droplets and a range of small molecules resulting from metabolic and catabolic processes. These have been utilized as indicators for numerous diseases such as lung ailments, breast cancer, diabetes, and other infectious conditions like influenza. The potential to expand their application to identify COVID-19 offers several advantages over conventional approaches \cite{Lamote2020, Song2020}.
Breath analysis is non-invasive, eliminating the need for a healthcare professional to collect samples using swabs or other invasive procedures, thereby avoiding any discomfort for the patient being tested. It enables rapid testing, ensuring prompt receipt of test results for early detection and the prevention of virus transmission. Furthermore, breath analysis is a relatively cost-effective and portable testing method, making it suitable not only for hospitals and clinics but also for locations with high population density, such as airports, where real-time testing can be of significant benefit.



\section*{Method}

We propose a detection system that leverages mass spectrometry and Artificial Intelligence (AI) to rapidly assess exhaled breath samples from patients and identify the presence of COVID-19. The approach eliminates the need for prior identification of specific volatile organic compounds (VOCs) and is based on the direct analysis of the mass spectrum. Breath samples can be conveniently stored in specialized containers, simplifying collection procedures that can be performed by non-specialized personnel in various locations.

Our system utilizes a proprietary nano-sampling device coupled with a high-precision mass spectrometer capable of performing real-time mass spectrum analysis within the \qtyrange{10}{351}{m/z} range; this analysis requires usually few seconds, and never more than few minutes. The raw data are processed by in-house developed tools: first they are aligned to the baseline, then filtered to reduce measurement noise, and eventually a process of data augmentation enhances the robustness and diversity. Eventually, standard Machine Learning (ML) classifiers are employed to detect the presence of COVID-19. The system operates without the need for reagents and generates no hazardous waste.

\subsection*{Breath Samples Collection}

The subject's breath is collected into a sampling tedlar bag with a defined volume of 3 liters, by having the subject blow through a straw directly into the bag. Then, the bag is connected to the inlet valve of the MS apparatus. The inlet valve can have two possible settings: the first setting allows for the sample mixture, at atmospheric pressure, to flow from the bag to the ionization chamber, directly through an original Micro Electro-Mechanical System (MEMS) interface; the second setting connects the MEMS interface to a membrane pump, in order to clean the inlet line, bringing it to vacuum conditions ($\simeq \qty{1e-3}{\milli\bar}$).


Mass spectra are recorded via a \emph{Varian 1200L} mass analyzer software, which allows the setting of some acquisition parameters like mass range, acquisition time, and electron multiplier (EM) voltage. The latter parameter ultimately sets the detector amplification factor. We recorded mass spectra in the following ranges:

\begin{itemize}
    \item \qtyrange{10}{51}{m/z}, with an acquisition time of \qty{10}{s} and EM voltage of \qty{1000}{V};
    \item \qtyrange{49}{151}{m/z}, with an acquisition time of \qty{14}{s} and EM voltage of \qty{1800}{V};
    \item \qtyrange{149}{251}{m/z}, with an acquisition time of \qty{14}{s} and EM voltage of \qty{1800}{V};
    \item \qtyrange{249}{351}{m/z}, with an acquisition time of \qty{14}{s} and EM voltage of \qty{1800}{V};
\end{itemize}

To avoid signal saturation, the amplification in the first mass range was reduced due to the presence of the most abundant breath components, namely $\text{CO}_2$ (\qty{44}{m/z}), $\text{N}_2$ (\qty{28}{m/z}) and $\text{O}_2$ (\qty{32}{m/z}). For each breath sample, from 10 to 20 acquisitions were taken.

\begin{figure}[h!]
    \centering
    \includegraphics[width=0.4\linewidth]{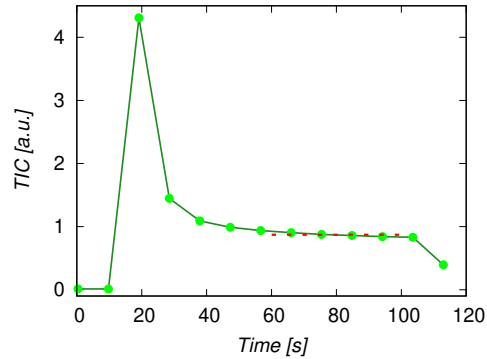}
    \caption{Total Ion Current (TIC) of a recording from one sample. The recording is made up of about 10 acquisitions (green dots), each corresponding to a mass spectrum. The spectra used for the analysis are selected on the plateau of TIC (red dotted region).}
    \label{fig:tic}
\end{figure}

By summing all the intensities for each \unit{m/z} in each acquisition, we can obtain the Total Ion Current (TIC) curve plot. Figure \ref{fig:tic} shows the TIC behavior when the breath sample flows into the MS system: the initial increase is due to the abrupt pressure change at the valve opening and, after a few tens of seconds, the TIC curve reaches a plateau region \cite{Correale2021}, when the flux stabilizes.

These procedures allowed for storing a dataset composed of the acquisitions of the spectra for each patient.

\subsection*{Pre-processing}
\label{sec:pre-processing}

Once the raw measures have been obtained, data are cleaned through a pre-processing procedure that reduces noise and machine variation of the acquisitions.

\begin{figure}
    \centering
    \includegraphics[width=0.7\linewidth]{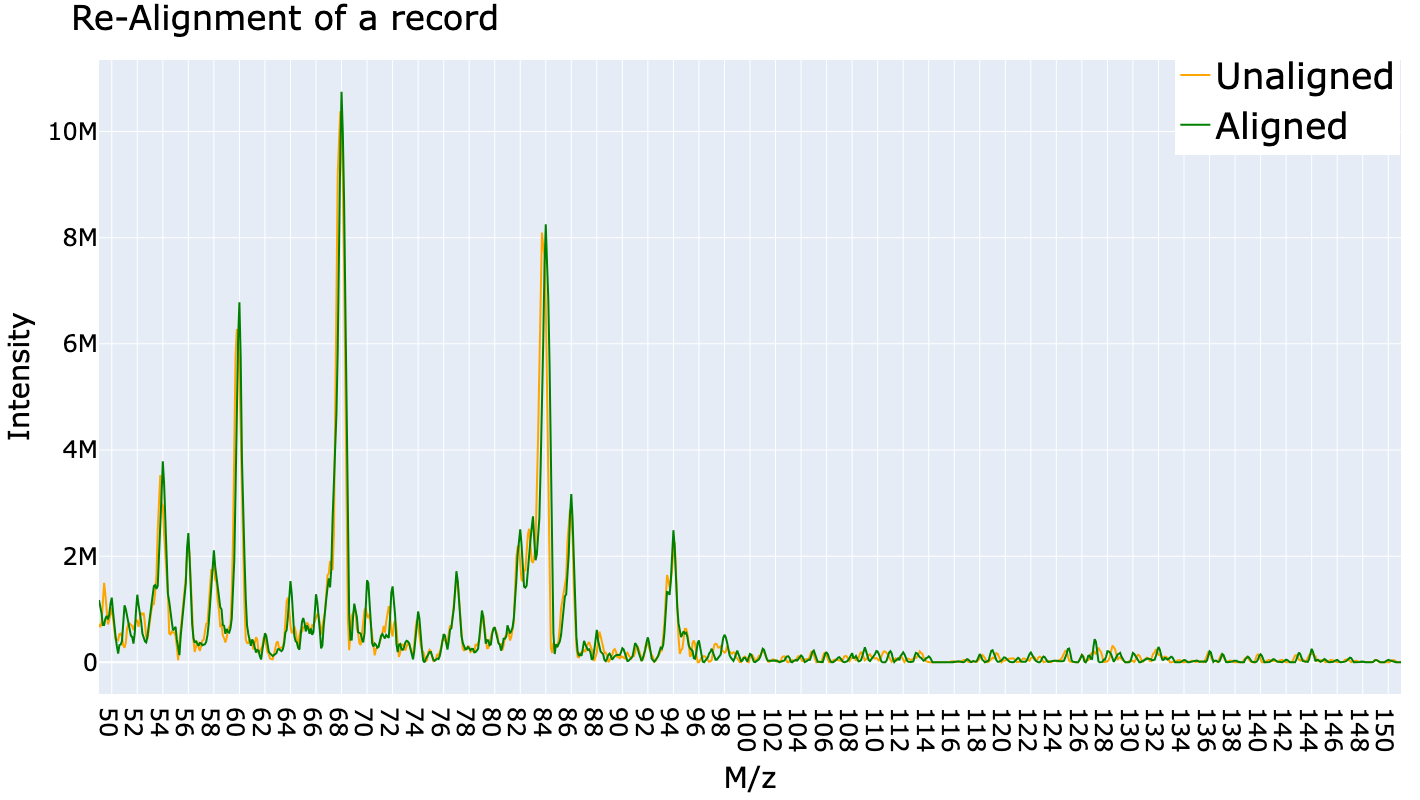}
    \caption{Aligned and non-aligned peaks of the mass spectrum of a single patient.}
    \label{fig:peak-alignment}
\end{figure}

For each acquisition, the recorded \unit{m/z} positions may be shifted with a specific alignment when the machine records the quantity of the ionized molecules due to measurement noise. A peak-alignment procedure is thus necessary. This procedure enables reducing the noise of the machine and compacting information. The peak alignment procedure is based on moving the peak to the nearest integer position, using them as anchors. The curves between two nearby peaks are stretched or compressed to sustain their original shape, preventing information loss. Peaks with mass $i$ are thus moved to the nearest integer position. Stretching and compression between peaks are done by linear interpolation to fit the corresponding segments in the reference. A graphical plot after the peak alignment can be seen in \cref{fig:peak-alignment}.

For each patient, we can obtain a single robust mass-spectra measurement by averaging the acquisitions in the plateau zone. To achieve this, first, it is necessary to identify the plateau in the TIC curve. A plateau-searching procedure was implemented, based on looking for acquisitions that do not vary so much from each other, by means of gradient computation of the signal. The plateau searching procedure was implemented as follows:

\begin{itemize}
    \item For each acquisition, we computed the gradient of the signals.
    \item A plateau is a zone that is almost flat, ideally where the gradient is zero, or in which the gradient does not vary so much from zero: we computed a tolerance guard-band $\epsilon$, to identify as ``flat'' a zone in which the gradient is near to zero (i.e., its absolute value is below $\epsilon$).
          This $\epsilon$ was computed on the basis of the $q-th$ quantile, in which $q$ is a number in the range $ [0,1] $ that indicates how tolerant the requirement of the constant slope of the plateau is.
    \item As shown in \cref{fig:tic}, the TIC curve may present more than one plateau: the first one is in the region in which the breath sample has not flown yet in the MS machine. This can be composed of 1 to 3 acquisitions. Thus, to avoid potential error, we considered the plateau of maximum length, that is in the region in which the ion flow stabilizes.
    \item Once the plateau of maximum length is found (that varies from 3 to 5 acquisitions, for each patient), we computed standard deviations of acquisitions in this region, by deploying a rolling window of size 4. We then chose the 4 acquisitions that minimize the standard deviation, and we computed the mean among these, obtaining a single spectrum for each patient.
\end{itemize}

Computing the average of the 4 acquisitions with minimum standard deviation permits the extraction of a single robust spectrum for each patient.

As an alternative, for each patient, we can insert all obtained acquisitions. This permits increasing both the number of training samples (by a factor of 4) and the variability in the data, thus, leading to more accurate models. During the testing phase, instead, we averaged the acquisitions, to have a single spectrum for each tested patient.

Some samples may present high noise in the mass spectrum: we considered outliers samples that have a $z$-score greater than 8, for at least one feature. For the same reason, for some patients was not possible to identify a plateau, and these are discarded from the dataset.

To overcome both noise in the measurements and possible parameter variations in the machine's setting, a signal smoothing procedure was implemented in the remaining patients. We first normalized each spectrum by dividing for the TIC value, to obtain relative information about the breath composition (i.e., each intensity was divided by the sum of all the intensities, thus scaling the feature in the range $(0,1)$). We then applied a high-pass filter, considering as noise (and thus, treating as zero) each intensity below \num{0.0001}. Then, we applied a Savitzky--Golay Smoothing and Differentiation Filter \cite{Savitzky_Golay_1964, Gallagher2020_SGfilter} to remove noise and align the signals to the baseline. This type of filter is used as a pre-processing step in spectra analysis to reduce both high-frequency noises, due to its smoothing properties, and low-frequency signals using differentiation. We then applied again the high-pass filter, considering as zero each intensity below \num{0.001}, to remove possible artifacts introduced by the filtering procedure.

The filtering and pre-processing procedure has been applied to each mass range, separately. At this point, it is possible to combine the obtained spectra in the four mass range, to obtain a single, whole, spectrum in the range 10-351.

If we previously considered different acquisitions for each mass range, merging them means computing all the combinations of the different acquisitions for each range. In other words, we are augmenting the dataset. In this perspective, a mass-spectrum is actually a particular combination of the different acquisition of each mass range, for each patient. This procedure is something similar to creating \emph{artificial} patients, in which each of them varies for one of the four pieces of the spectra. An example of the obtained augmented dataset is shown in \cref{tab:augmented_dataset}.

We then normalized again the whole spectrum by dividing it by the total sum of the intensity, obtaining only relative information.

\begin{table}[h!]
    \centering
    \begin{tabular}{|c|c|c|c|c|}
        \cline { 2 - 5 } \multicolumn{1}{c|}{} & \multicolumn{4}{c|}{ Acquisitions }                                              \\
        \hline Patient-ID                      & Mass-Range 1                        & Mass-Range 2 & Mass-Range 3 & Mass-Range 4 \\
        \hline 1-AAAA                          & 1                                   & 1            & 1            & 1            \\
        \hline 1-AAAB                          & 1                                   & 1            & 1            & 2            \\
        \hline 1-AAAC                          & 1                                   & 1            & 1            & 3            \\
        \hline$\ldots$                         &                                     &              &              &              \\
        \hline 1-ABCD                          & 1                                   & 2            & 3            & 4            \\
        \hline$\ldots$                         &                                     &              &              &              \\
        \hline 1-DDDD                          & 4                                   & 4            & 4            & 4            \\
        \hline 2-AAAA                          & 1                                   & 1            & 1            & 1            \\
        \hline$\ldots$                         &                                     &              &              &              \\
        \hline
    \end{tabular}
    \caption{An example of the dataset augmentation procedure. Each row is a pseudo-patient, generated by a particular combination of the different mass-range acquisitions of each actual patient.}
    \label{tab:augmented_dataset}
\end{table}

\subsection*{Machine-Learning Models}
\label{sec:ml-models}

Several Machine-Learning (ML) techniques have been proposed in the context of COVID-19 \cite{Costa2022MLCovid, pmid24713999}. We propose to exploit a mixture of solid, state-of-the-art ML models: K-nearest neighbors (KNN), random forest (RF), logistic regression (LR), gradient boosting (xGB), support vector machine (SVM) with \textit{RBF} kernel, and an ensemble made by all the models, in a soft-voting fashion (Ens).
Soft voting is a computational method employed to amalgamate predictions from an ensemble of classifiers, leveraging the probabilities associated with each prediction. In this approach, each classifier initially assigns a probability to each class. The final prediction of the ensemble is determined by selecting the class with the highest cumulative probability across all classifiers involved in the voting process.

We tried several feature pre-processing techniques, that led to several ML models, whose results are presented in the next sections. First, all the features that present a variance equal to zero are pruned. In other words, we removed all the \unit{m/z} for which no intensities were measured after the filtering procedure.

Each feature is then individually normalized. We propose two different types of features normalization: the \textit{Standard Scaler}, and the \textit{Robust Scaler}
The \textit{Standard Scaler, SS} acts by subtracting the mean and scaling the value according to the variance.
The \textit{Robust Scaler, RS}, instead, subtracts the median and scales the value according to the quantile range between the \nth{1} quartile and the \nth{3} quartile. This type of scaling, sometimes referred to as \emph{robust}, reduces the influence of the outliers.

A further step of feature reduction is performed to select only the most informative features. We used Feature Selection, which means to select the features the most linked to the target task in a supervised feature selection fashion. For this purpose, we used SURF*, a Relief-based algorithm \cite{Kira1992_Relief}. We also tried a principal component analysis (PCA) \cite{gao2020principal,jolliffe2002principal} to extract the first 20 principal components of our dataset.

Then, the different pre-processing pipelines were applied to the obtained dataset, and the chosen ML models were trained to obtain classification systems to discriminate between COVID-19 positive and negative patients.

\section*{Experimental Evaluation}
\label{sec:experiments}
\subsection*{Samples Collection}

Breath samples were collected from patients and medical personnel at Varese Hospital (Ospedale di Circolo --- Fondazione Macchi, ASST Sette Laghi). The acquisition lasted one year, from March 2021 to March 2022, for a total of 302 tested subjects. Some patients have been tested more than once to calibrate the system. The mass spectra have been collected with a Varian 1200L mass analyzer, combined with the MEMS interface. As ground truth to confirm SARS-CoV-2 infection, a RT-qPCR nasopharyngeal swab testing has been performed on all subjects.

\begin{figure*}
    \centering
    \includegraphics[width=1\linewidth]{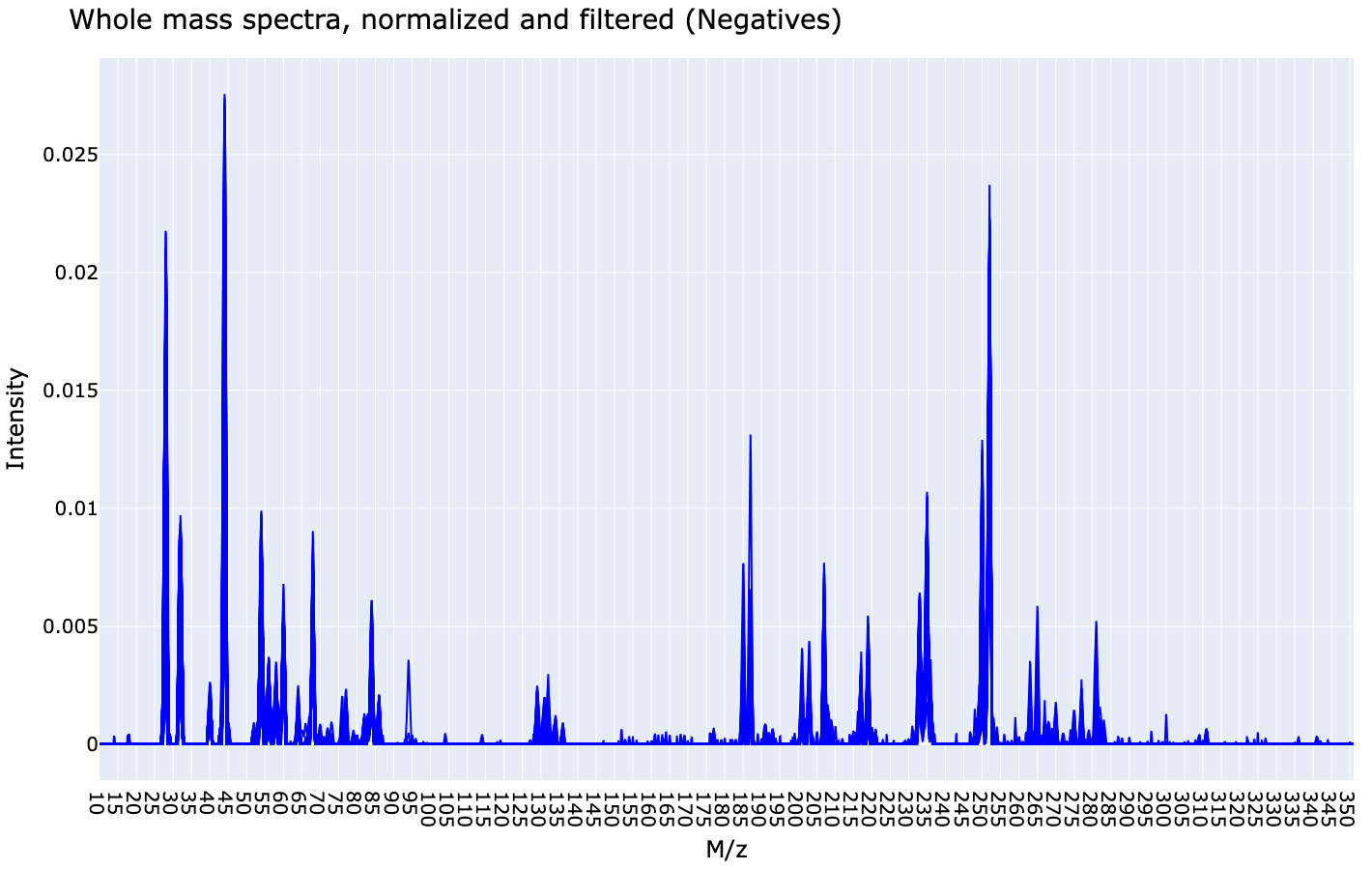} \\
    \includegraphics[width=1\linewidth]{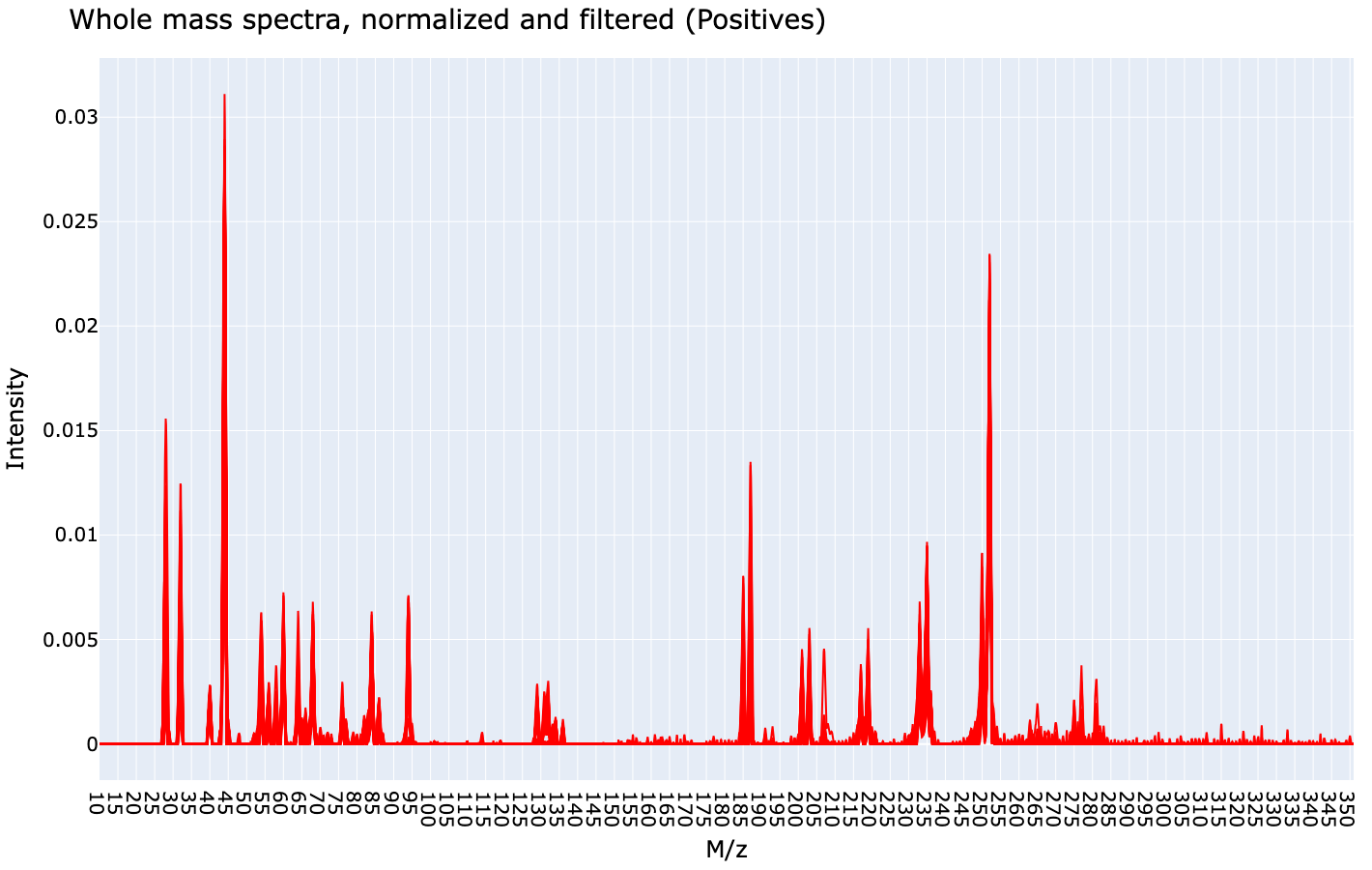}
    \caption{Examples of whole spectra (\qtyrange{10}{351}{m/z}) for negative subjects (top, blue) and positive ones (bottom, red)}
    \label{fig:whole_range_spectra}
\end{figure*}

The raw dataset presents breath samples for a total of \num{1208} acquisitions in 302 patients, divided into 91 positive and 211 negative records. After the outliers' removal procedures and plateau identification, were retained only 287 patients for mass-range 2 and 203 for the whole spectrum 10-351. These problems were caused by the highly prototypical nature of the equipment; it is worth noticing that, in a real application, it would have been possible to repeat the measurement.

\begin{figure*}
    \centering
    \minipage{0.5\linewidth}
    \includegraphics[width=1\linewidth]{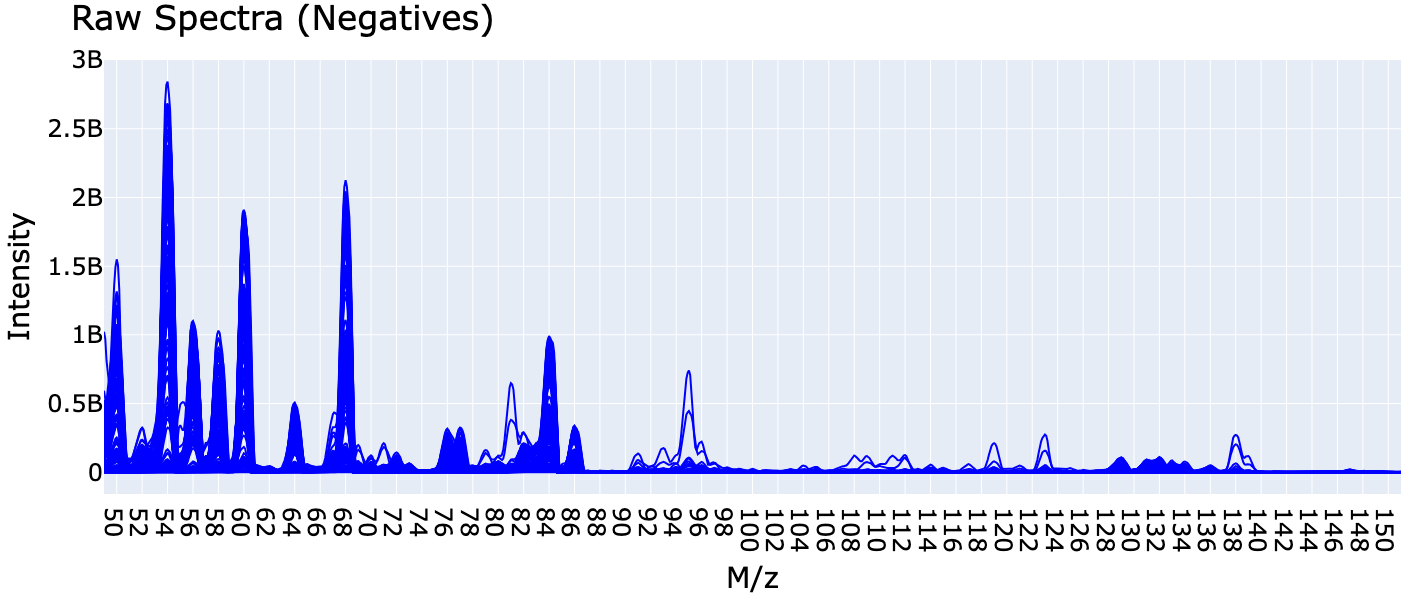}
    \endminipage\hfill
    \minipage{0.5\linewidth}
    \includegraphics[width=1\linewidth]{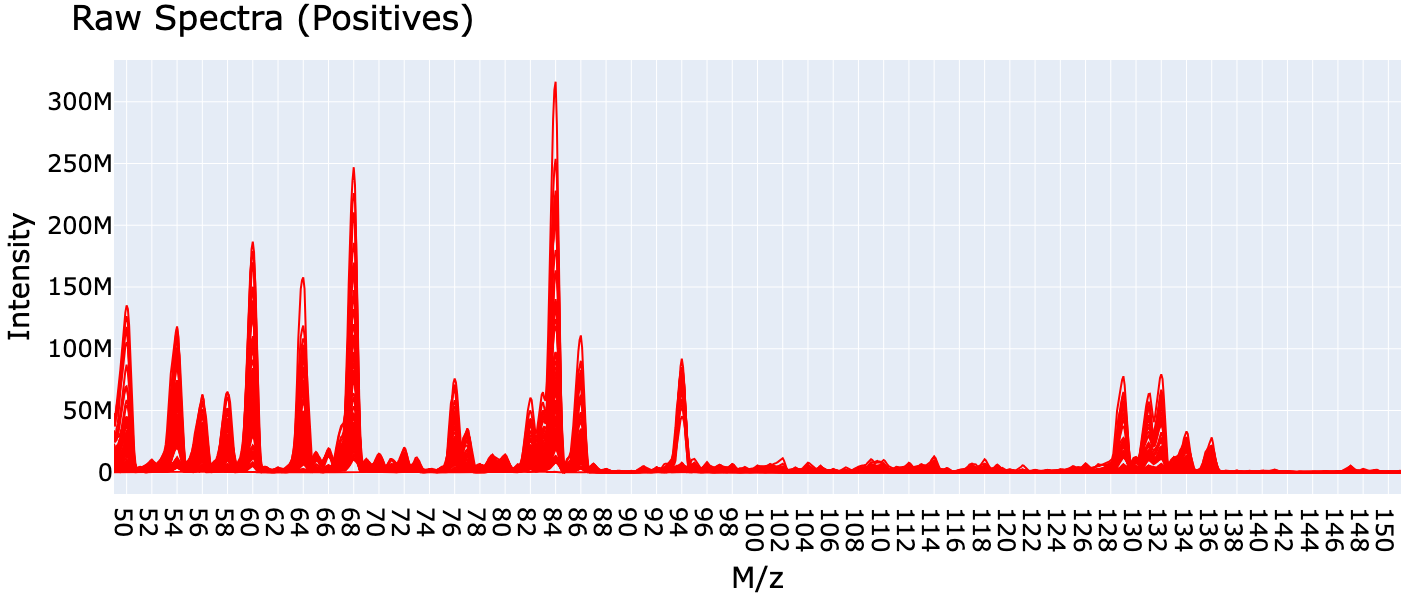}
    \endminipage\hfill

    \minipage{0.5\linewidth}
    \includegraphics[width=1\linewidth]{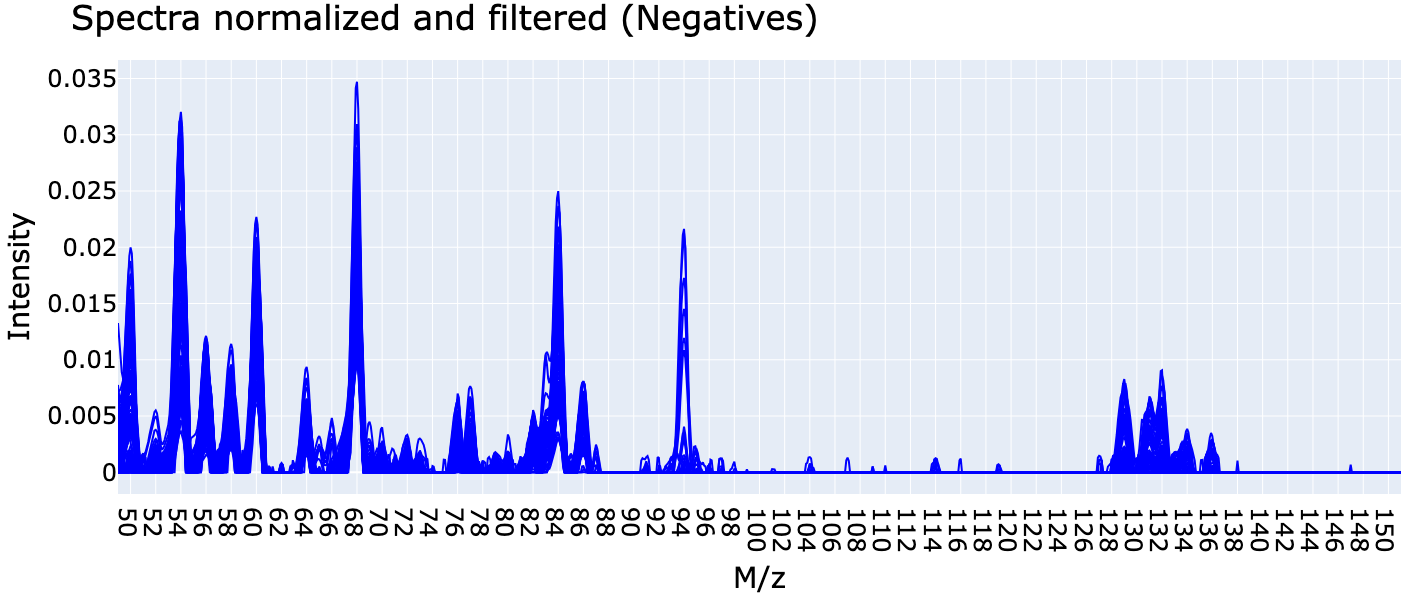}
    \endminipage\hfill
    \minipage{0.5\linewidth}
    \includegraphics[width=1\linewidth]{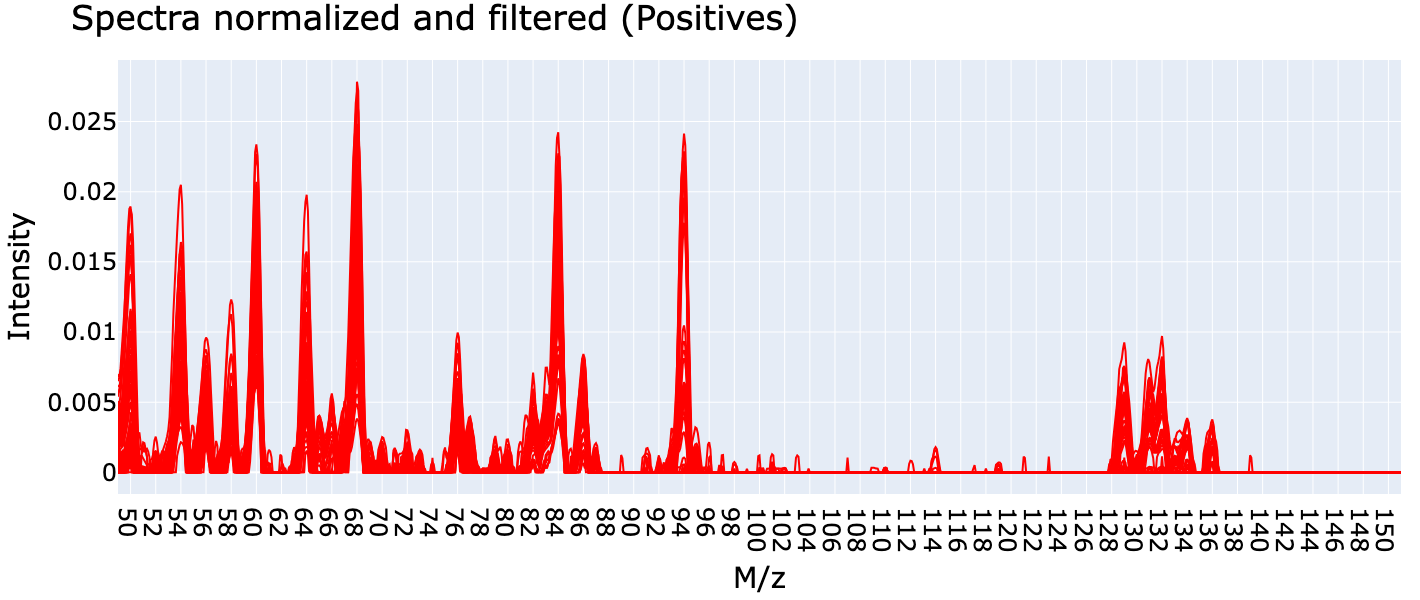}
    \endminipage\hfill

    \caption{The comparison of spectra before (top) and after the filtering and normalizing procedure (bottom) shows the removal of low-frequency noise. Negative patients are on the left (blue), and positive on the right (red).}
    \label{fig:filtered_positive_negative_range_2}
\end{figure*}

A graphical view of the results of the proposed pre-processing filter procedure described before can be found in \cref{fig:filtered_positive_negative_range_2} for mass-range 2 and in \cref{fig:whole_range_spectra} for the whole range under analysis. Filtering the spectra allows us to reduce instrument variability and noise.

\begin{figure}
    \centering
    \includegraphics[width=\linewidth]{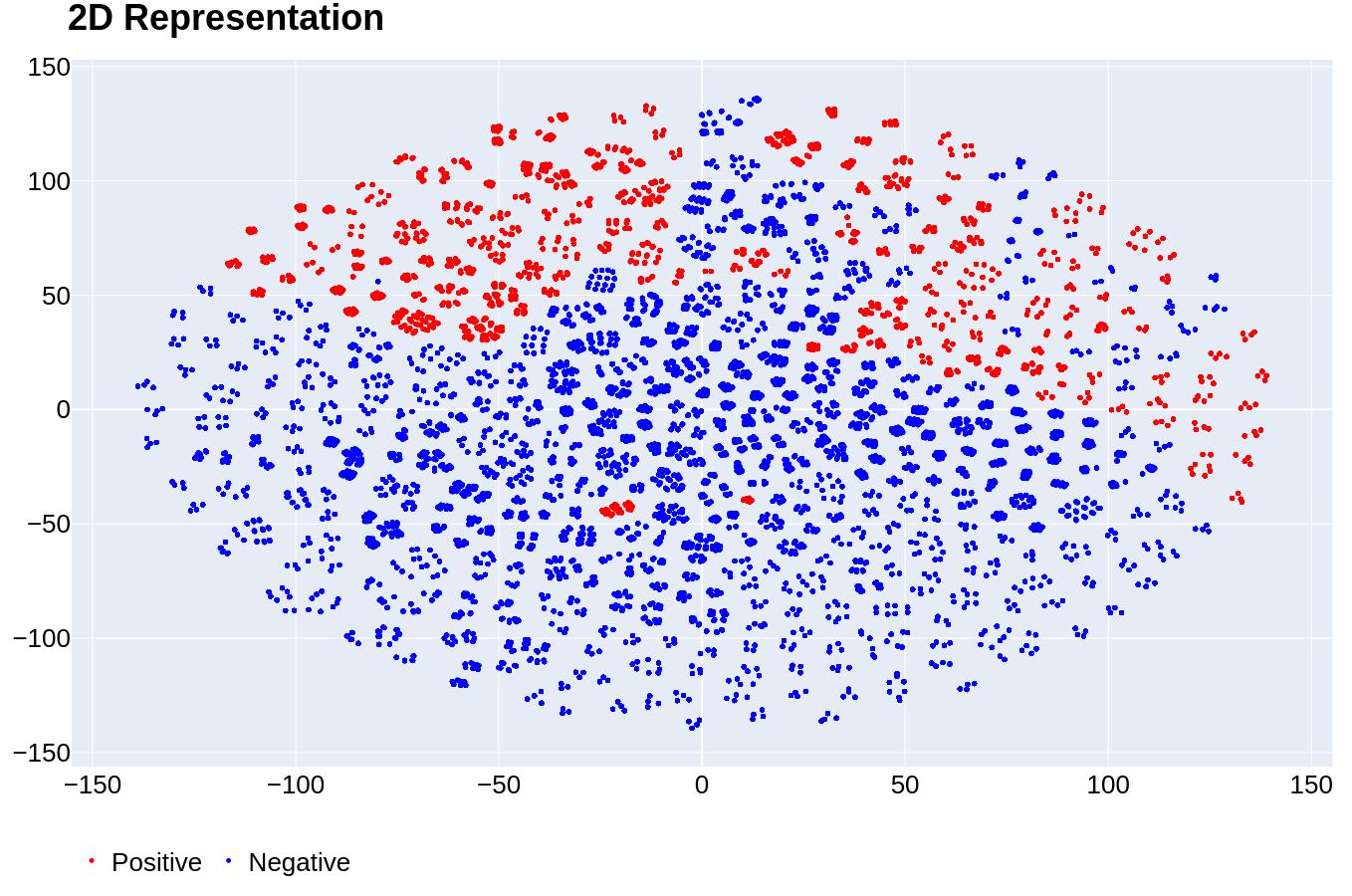}
    \caption{2D t-SNE representation of the whole spectra for all the 47,084 samples generated.}
    \label{fig:2d_whole_ranges}
\end{figure}

The data-augmentation procedure, by considering all the different combinations of the acquisitions, leads to the generation of \num{47084} samples. A graphical 2D representation of the points obtained with t-SNE dimensionality reduction \cite{JMLR:v9:vandermaaten08a} is presented in figure \ref{fig:2d_whole_ranges}. There, a sharp boundary between positive and negative samples can be seen.

\subsection*{Performance Evaluation}
\label{sec:evaluation}

Patients have been split into training and test data: the training data are used to create the ML models, while the testing data are used only for evaluation purposes. We evaluated the experiments on 10 different training-test splits, averaging the results to obtain an unbiased estimation of the generalization performances of our models. We used a 10-fold Stratified Cross Validation: the acquisitions of each patient are not mixed among training and test sets, to avoid potential information leakages.

To address the issue of the high class-imbalance, we utilized a simple oversampling technique of the minority class (i.e., the COVID-19-positive class) in training sets, obtaining the same number of positive and negative samples.

Results are presented in terms of famous classification performances: Balanced Accuracy, Precision, Recall, and $F_1$-score.

These metrics are computed based on the number of samples correctly and incorrectly predicted by our models.

Precision is the ability of the classifier not to label as positive a sample that is negative, while recall is the ability of the classifier to find all the positive samples.
The balanced accuracy avoids inflated performance estimates on imbalanced datasets. It is the macro-average of recall scores, for each class (the mean of recall for negative and positive classes).

The $F_1$-score is the harmonic mean of the precision and recall.
The relative contribution of precision and recall to the $F_1$-score are equal.
All of these metrics lies in the range $[0,1]$, where a score equal to 1 (or 100\%) means perfect classification performance.

In the experiments on the whole mass spectrum 10-351, with the four ranges together, we also computed two additional performance metrics: Specificity and the area under the receiver operating characteristic curve (ROC-AUC).
While recall is a measure that evaluates a test's ability to correctly identify unhealthy individuals, specificity carries the same concept but for healthy patients.
Specificity is the recall of the negative class.
When a test exhibits high specificity, a positive result becomes valuable in confirming the presence of the disease, as the test rarely produces positive outcomes in healthy individuals. 

A receiver operating characteristic (ROC) curve is a visual representation that showcases the performance of a binary classifier system as the threshold for classification is adjusted. It plots the true positive rate \textit{(TPR)} against the false positive rate \textit{(FPR}) at different threshold values. \textit{TPR} is also referred to as sensitivity, while \textit{FPR} is the complement of specificity.

The ROC-AUC quantifies the overall performance of the classifier by calculating the area under the ROC curve.

\section*{Results}

We first trained the models on the different mass ranges, separately and with no feature pre-processing or pruning.
The first experiments indicated that mass range 2 was the more useful for classification: the Ensemble models show a $F_1$-score of 71\% and accuracy of 84\%.(\cref{tab:different_ranges}).
Then we applied the filtering and normalization procedures and the features pre-processing and pruning on Mass Range 2. The results of the different trials can be seen in \cref{tab:merged_table}.
Processing the spectra and the features is beneficial, permitting an increase in the classification performances for all the models taken into account. As an example, we can compare the performances of the Ensemble model in \cref{tab:different_ranges,tab:merged_table}: it is evident that, for range 2, we increased the prediction performances (Accuracy increased from 84\% to 87\%, and $F_1$-Score from 70\% to about 80\%).

Also, the idea of inserting multiple acquisitions for each patient in the training set, while averaging them on the test set, led to a lower prediction error and higher classification performance with respect to a single, robust spectrum (\cref{tab:merged_table}).
As a claim of this, we can consider the SVC model: with a single acquisition per patient, we have 76\% Accuracy and only 57\% of Recall, meaning that our models is not accurate in detecting positive patients. If we consider multiple acquisitions, instead, we have 89\% of Accuracy and 91\% of Recall.

Including a Feature Selection step was not beneficial regarding the error metrics: with the SURF* algorithm, the accuracy metrics decreased. It may be due to complex interaction between features, that this algorithm is not able to catch. The Robust Scaler (RS, in \cref{tab:merged_table}) over the Standard Scaler (SS) permits decreasing the standard deviation in the performance metrics, and thus the error fluctuations, due to the ability to deal with outliers values. It is extremely beneficial in the case of SVC.

In general, SVC and Ensemble models have the highest recall. They hold the highest ability to correctly identify true positive patients.

In principle, prediction performances could be additionally improved with a fine search in the hyperparameters space.


The pre-processing steps for the whole mass range experiments (10-351, results in \cref{tab:all_ranges}) included the data augmentation procedure, the spectra normalization and filtering, the robust scaler, and the PCA feature reduction: these steps have been recognized as the ones that permitted the best accuracy on mass-range 2 experiments.

This setup permitted used to achieve the best performances in terms of error metric \cref{tab:all_ranges}: with the ensemble method, we achieve 95\% accuracy, 94\% recall, 96\% specificity, and $F_1$-score of 92\%.

We recall that for this last strategy, the dataset set is composed of about 47,000 samples, but the patients under testing are 203.

\begin{table}[h!]
    \centering
    \caption{Mean Results on test sets (10 splits) with different mass ranges. No features pre-processing}
    \resizebox{0.5\linewidth}{!}{%
        \begin{tabular}{cccccc}
            \toprule
            \textbf{Model} & \textbf{Mass Range} & \textbf{Accuracy} & \textbf{Precision} & \textbf{Recall} & \textbf{$F_1$-Score} \\
            \midrule
            RF             & 1                   & 0.78              & 0.65               & 0.56            & 0.60                 \\
            RF             & 2                   & 0.82              & 0.69               & 0.69            & 0.68                 \\
            RF             & 3                   & 0.82              & 0.72               & 0.57            & 0.63                 \\
            RF             & 4                   & 0.79              & 0.65               & 0.53            & 0.57                 \\
            \hline
            Ens            & 1                   & 0.82              & 0.70               & 0.67            & 0.68                 \\
            Ens            & 2                   & 0.84              & 0.75               & 0.69            & 0.71                 \\
            Ens            & 3                   & 0.80              & 0.68               & 0.57            & 0.61                 \\
            Ens            & 4                   & 0.77              & 0.62               & 0.60            & 0.59                 \\
            \bottomrule
        \end{tabular}}
    \label{tab:different_ranges}
\end{table}

\begin{table}[h]
    \centering
    \caption{Mean Results on tes sets (10 splits) on mass range 2}
    \label{tab:merged_table}
    \resizebox{\linewidth}{!}{%
        \begin{tabular}{lccccccc}
            \hline
            \textbf{Alg.} & \textbf{Filtering} & \textbf{Feat. Sel.} & \textbf{Acquisition} & \textbf{Accuracy}        & \textbf{Precision}       & \textbf{Recall}          & \textbf{$F_1$-Score}     \\
            \hline
            xGB           & No                 & PCA                 & Single               & 0.83 $\pm$ 0.04          & 0.74 $\pm$ 0.10          & 0.77 $\pm$ 0.08          & 0.75 $\pm$ 0.06          \\
            xGB           & Yes (SS)           & PCA                 & Single               & 0.88 $\pm$ 0.07          & 0.79 $\pm$ 0.08          & 0.86 $\pm$ 0.13          & 0.82 $\pm$ 0.09          \\
            xGB           & Yes (SS)           & PCA                 & Multiple             & 0.93 $\pm$ 0.05          & 0.85 $\pm$ 0.14          & 0.94 $\pm$ 0.07          & 0.88 $\pm$ 0.09          \\
            xGB           & Yes (SS)           & SURF*, PCA          & Multiple             & 0.86 $\pm$ 0.07          & 0.78 $\pm$ 0.14          & 0.83 $\pm$ 0.10          & 0.80 $\pm$ 0.10          \\
            xGB           & Yes (RS)           & PCA                 & Multiple             & 0.90 $\pm$ 0.04          & 0.82 $\pm$ 0.10          & 0.89 $\pm$ 0.08          & 0.84 $\pm$ 0.05          \\
            \hline
            KNN           & No                 & PCA                 & Single               & 0.89 $\pm$ 0.05          & 0.73 $\pm$ 0.09          & 0.92 $\pm$ 0.08          & 0.81 $\pm$ 0.07          \\
            KNN           & Yes (SS)           & PCA                 & Single               & 0.87 $\pm$ 0.05          & 0.73 $\pm$ 0.07          & 0.89 $\pm$ 0.08          & 0.80 $\pm$ 0.07          \\
            KNN           & Yes (SS)           & PCA                 & Multiple             & 0.91 $\pm$ 0.07          & 0.80 $\pm$ 0.14          & 0.93 $\pm$ 0.10          & 0.85 $\pm$ 0.11          \\
            KNN           & Yes (SS)           & SURF*, PCA          & Multiple             & 0.85 $\pm$ 0.07          & 0.70 $\pm$ 0.13          & 0.87 $\pm$ 0.10          & 0.77 $\pm$ 0.11          \\
            KNN           & Yes (RS)           & PCA                 & Multiple             & 0.91 $\pm$ 0.05          & 0.78 $\pm$ 0.12          & 0.94 $\pm$ 0.08          & 0.84 $\pm$ 0.08          \\
            \hline
            LR            & No                 & PCA                 & Single               & 0.86 $\pm$ 0.05          & 0.71 $\pm$ 0.12          & 0.88 $\pm$ 0.07          & 0.78 $\pm$ 0.08          \\
            LR            & Yes (SS)           & PCA                 & Single               & 0.85 $\pm$ 0.06          & 0.71 $\pm$ 0.08          & 0.85 $\pm$ 0.10          & 0.77 $\pm$ 0.07          \\
            LR            & Yes (SS)           & PCA                 & Multiple             & 0.88 $\pm$ 0.06          & 0.73 $\pm$ 0.14          & 0.92 $\pm$ 0.10          & 0.80 $\pm$ 0.10          \\
            LR            & Yes (SS)           & SURF*, PCA          & Multiple             & 0.88 $\pm$ 0.07          & 0.74 $\pm$ 0.13          & 0.89 $\pm$ 0.10          & 0.80 $\pm$ 0.10          \\
            LR            & Yes (RS)           & PCA                 & Multiple             & 0.89 $\pm$ 0.05          & 0.76 $\pm$ 0.11          & 0.91 $\pm$ 0.08          & 0.82 $\pm$ 0.08          \\
            \hline
            RF            & No                 & PCA                 & Single               & 0.87 $\pm$ 0.05          & 0.79 $\pm$ 0.10          & 0.84 $\pm$ 0.09          & 0.81 $\pm$ 0.07          \\
            RF            & Yes (SS)           & PCA                 & Single               & 0.89 $\pm$ 0.04          & 0.82 $\pm$ 0.11          & 0.87 $\pm$ 0.08          & 0.84 $\pm$ 0.05          \\
            RF            & Yes (SS)           & PCA                 & Multiple             & 0.91 $\pm$ 0.07          & 0.80 $\pm$ 0.14          & 0.92 $\pm$ 0.09          & 0.85 $\pm$ 0.10          \\
            RF            & Yes (SS)           & SURF*, PCA          & Multiple             & 0.85 $\pm$ 0.09          & 0.75 $\pm$ 0.16          & 0.82 $\pm$ 0.13          & 0.78 $\pm$ 0.12          \\
            RF            & Yes (RS)           & PCA                 & Multiple             & 0.90 $\pm$ 0.03          & 0.81 $\pm$ 0.07          & 0.90 $\pm$ 0.07          & 0.84 $\pm$ 0.04          \\
            \hline
            SVC           & No                 & PCA                 & Single               & 0.65 $\pm$ 0.07          & 0.82 $\pm$ 0.20          & 0.32 $\pm$ 0.13          & 0.45 $\pm$ 0.16          \\
            SVC           & Yes (SS)           & PCA                 & Single               & 0.76 $\pm$ 0.10          & 0.82 $\pm$ 0.15          & 0.57 $\pm$ 0.17          & 0.66 $\pm$ 0.15          \\

            SVC           & Yes (SS)           & PCA                 & Multiple             & 0.89 $\pm$ 0.05          & 0.76 $\pm$ 0.13          & 0.91 $\pm$ 0.13          & 0.81 $\pm$ 0.07          \\

            SVC           & Yes (SS)           & SURF*, PCA          & Multiple             & 0.86 $\pm$ 0.08          & 0.77 $\pm$ 0.17          & 0.83 $\pm$ 0.09          & 0.79 $\pm$ 0.12          \\
            \textbf{SVC}  & \textbf{Yes (RS)}  & \textbf{PCA}        & \textbf{Multiple}    & \textbf{0.93 $\pm$ 0.04} & \textbf{0.85 $\pm$ 0.09} & \textbf{0.94 $\pm$ 0.07} & \textbf{0.89 $\pm$ 0.06} \\
            \hline
            \hline
            Ens.          & No                 & PCA                 & Single               & 0.87 $\pm$ 0.03          & 0.77 $\pm$ 0.10          & 0.84 $\pm$ 0.06          & 0.80 $\pm$ 0.06          \\
            Ens.          & Yes (SS)           & PCA                 & Single               & 0.90 $\pm$ 0.07          & 0.82 $\pm$ 0.11          & 0.89 $\pm$ 0.10          & 0.85 $\pm$ 0.09          \\
            Ens.          & Yes (SS)           & PCA                 & Multiple             & 0.93 $\pm$ 0.07          & 0.83 $\pm$ 0.15          & 0.94 $\pm$ 0.09          & 0.87 $\pm$ 0.11          \\
            Ens.          & Yes (SS)           & SURF*, PCA          & Multiple             & 0.89 $\pm$ 0.07          & 0.79 $\pm$ 0.16          & 0.88 $\pm$ 0.08          & 0.82 $\pm$ 0.12          \\
            Ens.          & Yes (RS)           & PCA                 & Multiple             & 0.92 $\pm$ 0.04          & 0.81 $\pm$ 0.09          & 0.94 $\pm$ 0.07          & 0.87 $\pm$ 0.06          \\
            \hline
        \end{tabular}}
\end{table}

\begin{table*}[htbp]
    \centering
    \caption{Mean Results on the test sets (10 splits) for the whole mass range 10-351. Filtering the spectra and pre-processing the features (Robust Scaler and PCA) were applied}
    \resizebox{0.8\linewidth}{!}{%
        \begin{tabular}{lcccccc}
            \toprule
            \textbf{Alg.} & \textbf{Accuracy}        & \textbf{Precision}        & \textbf{Recall}          & \textbf{$F_1$-Score}     & \textbf{Specificity}     & \textbf{ROC-AUC}         \\
            \midrule
            KNN           & 0.93 $\pm$ 0.06          & 0.87 $\pm$ 0.09           & 0.92 $\pm$ 0.09          & 0.89 $\pm$ 0.08          & 0.94 $\pm$ 0.04          & 0.95 $\pm$ 0.04          \\
            RF            & 0.91 $\pm$ 0.06          & 0.88 $\pm$ 0.10           & 0.87 $\pm$ 0.12          & 0.87 $\pm$ 0.07          & 0.95 $\pm$ 0.04          & 0.98 $\pm$ 0.03          \\
            LR            & 0.94 $\pm$ 0.04          & 0.84 $\pm$ 0.12           & 0.96 $\pm$ 0.07          & 0.89 $\pm$ 0.07          & 0.93 $\pm$ 0.05          & 0.97 $\pm$ 0.04          \\
            xGB           & 0.94 $\pm$ 0.03          & 0.88 $\pm$ 0.08           & 0.93 $\pm$ 0.07          & 0.90 $\pm$ 0.03          & 0.95 $\pm$ 0.03          & 0.98 $\pm$ 0.03          \\
            SVC           & 0.93 $\pm$ 0.06          & 0.89 $\pm$ 0.09           & 0.90 $\pm$ 0.12          & 0.88 $\pm$ 0.06          & 0.95 $\pm$ 0.04          & 0.98 $\pm$ 0.02          \\
            \hline
            \textbf{Ens.} & \textbf{0.95 $\pm$ 0.04} & \textbf{ 0.90 $\pm$ 0.08} & \textbf{0.94 $\pm$ 0.07} & \textbf{0.92 $\pm$ 0.05} & \textbf{0.96 $\pm$ 0.03} & \textbf{0.98 $\pm$ 0.03} \\
            \bottomrule
        \end{tabular}}
    \label{tab:all_ranges}
\end{table*}

\section*{Discussion}
\label{sec:conclusions}
We presented a framework for COVID-19 detection by breath samples. We used a prototype of a special MS portable machine based on nanotechnology to analyze human breath in about 2 minutes, extracting the mass spectrum of a patient in the interval 10-351, divided into four sub-ranges. Experiments showed that the mass spectra can be related to the presence of COVID-19 using ML classification models. We proposed a filtering procedure based on Savitzky-Golay filter to reduce possible noise in the acquisitions. Results showed that with simple approaches we are able to get about 93\% accuracy and 94\% recall with mass range of \qtyrange{49}{151}{m/z}, which seems to be the most appropriate in predicting COVID-19 disease. We found out that the use of Robust scaling techniques, based on the median and IQR scaling, in conjunction with PCA feature extraction was beneficial in predicting COVID-19 diseases. We were able to merge all the spectra ranges, reaching classification performances of about 95\% accuracy, 94\% recall, and 98\% ROC-AUC score. In general, the special portable MS machine is potentially deployable in COVID-19 hubs to detect in a fast, easy, comfortable way the presence or not of COVID-19. The use of ML to relate mass spectra to special diseases enables early detection, rapid test results, and less risk of infection for healthcare providers.

\section*{Acknowledgement}
The authors wish to thank Matteo Serra for his contribution.

\bibliographystyle{unsrtnat}
\bibliography{references.bib}

\end{document}